\documentclass{article}
\usepackage{spconf,amsmath,graphicx,bm,accents}
\usepackage{amsfonts}       
\newcommand{\minimize}{{\rm minimize}}

\usepackage{enumitem,kantlipsum}
\usepackage{algorithm}
\usepackage{algorithmic}

\title{dPCA: Dimensionality Reduction for Discriminative
	 Analytics\\ of Multiple Large-Scale Datasets}

\name{Gang Wang, Jia Chen, and Georgios B. Giannakis\thanks{Work in this paper was supported in part by NIH 1R01GM104975-01 and NSF 1500713.}}
\address{ECE Dept. and Digital Tech. Center, U. of Minnesota, Mpls., MN 55455, USA\\ Emails: \{gangwang, chen5625, georgios\}@umn.edu}
\begin{document}
%
\maketitle
\begin{abstract}
Principal component analysis (PCA) has well-documented merits for data extraction and dimensionality reduction. PCA deals with a single dataset at a time, and it is challenged when it comes to analyzing multiple datasets. Yet in certain setups, one wishes to extract the most significant information of one dataset relative to other datasets. Specifically, the interest may be on identifying, namely extracting features that are specific to a single target dataset but not the others. This paper develops a novel approach for such so-termed \emph{discriminative data analysis}, and establishes its \emph{optimality} in the least-squares (LS) sense under suitable data modeling assumptions. The criterion reveals linear combinations of variables by maximizing the ratio of the variance of the target data to that of the remainders. The novel approach solves a generalized eigenvalue problem by performing SVD just once. Numerical tests using synthetic and real datasets showcase the merits of the proposed approach relative to its competing alternatives.  
\end{abstract}

%

\section{Introduction}
\label{sec:intro}

Principal component analysis (PCA) is arguably the most widely used method for data visualization and dimensionality reduction \cite{2010pca}. 
PCA originated in statistics
\cite{1901pca}, 
but its modern instantiation as well as the term \emph{principal component} (PC) vector was formalized
in \cite{1933pca}. The goal of PCA is to extract the most important information from a data table representing observations, and depict it as a few PCs. PCs are 
uncorrelated 
linear transformations of the original set of variables, along which the 
maximum variation in the data is captured 
\cite{2010pca}. 


Yet, several application domains involve \emph{multiple} datasets, and the task is to extract
 trends or
  features depicted by \emph{component vectors} that are present in
one dataset \emph{but not} the other(s) \cite{2017cpca}. 
 For example, consider two gene expression datasets of individuals from across different countries and genders: the first includes gene expression levels of cancer patients, which constitutes the \emph{target data} 
 that we want to analyze, while the second is formed by healthy volunteers, and is called \emph{control} or \emph{background data}. 
Applying PCA on either the target data 
 or the target augmented with the control data
  is likely to obtain
   PCs that 
  represent
   the background information 
common to
 both datasets (e.g., the demographic patterns, genders) \cite{1998background}, 
instead of the component vectors depicting the subtypes of cancer within patients. 
Despite 
its practical
 relevance, 
 such discriminative data analysis has not been widely studied. 

Generalizations to PCA include multi-dimensional scaling \cite{mds},
locally linear embedding \cite{lle}, sparse or kernel PCA \cite{kpca}, \cite{spca}, and canonical correlation analysis \cite{1936cca}, \cite{tsp2016cs}, \cite{sp2016cs}, \cite{sp2017cs}. 
Given multiple datasets, 
analysts have to perform these 
procedures on each individual dataset, and subsequently evaluate manually the obtained projections to identify whether significant patterns representing similarities or differences across datasets are present.  
A recent approach however, pursued what is termed
contrastive (c) PCA for extracting the most distinct features of one dataset relative to the other \cite{2017cpca}. cPCA is able to reveal the dataset-specific information often
missed by PCA 
\cite{2017cpca}. This becomes possible through a hyper-parameter that takes into account both target and control data, and critically affects the performance of cPCA. The resultant cPCA algorithm requires finding the top eigenvectors of 
 indefinite matrices, that is often done with SVD. 
 Although 
 possible to automatically select the best from a list of given values, computing SVD multiple times 
 can be computationally cumbersome or even prohibitive in large-scale data extraction settings.  
Another method  related to PCA is linear discriminant analysis (LDA) \cite{1933lda}, 
that is ``supervised,'' and seeks linear combinations of variables to maximize the separation between classes.
This is achieved by maximizing the ratio of the variance between classes to the variance within the classes. 

Inspired by LDA and cPCA, this paper puts forth a new method for discriminative analytics, which is shown to be optimal in the least-squares (LS) sense provided that the background component vectors are also present in the target data.
Our method seeks linear combinations of variables by maximizing the ratio of the variance of target data to the variance of  control data, which justifies our chosen description as \emph{discriminative (d) PCA}. 
dPCA solves a generalized eigenvalue problem.
Relative to cPCA, dPCA is parameter-free, and requires only one SVD. As such, dPCA is well-suited for
large-scale either discriminative or contrasting data exploration. 


\section{Preliminaries and Prior Art}
\label{sec:ps}

Consider two datasets, the target data $\{\bm{x}_i\in\mathbb{R}^D \}_{1\le i\le m}$ that we are interested in analyzing, and data $\{\bm{y}_j \in\mathbb{R}^D\}_{1\le j\le n}$ containing latent background component vectors in the target data. Assume without loss of generality (wlog) that the sample mean of each dataset has been removed, and let $\bm{C}_{xx}:=(1/m)\sum_{i=1}^m\bm{x}_i\bm{x}_i^\top$ and $\bm{C}_{yy}:=(1/n)\sum_{i=1}^n\bm{y}_i\bm{y}_i^\top$ denote the corresponding sample covariance matrices. To motivate the novel approach in Sec. \ref{sec:dpca}, the basics of PCA and cPCA are briefly reviewed in this section.


 One formulation of PCA seeks vectors $\{\boldsymbol{\chi}_i\in\mathbb{R}^d \}_{1\le i\le m}$ as linear combinations of $\{\bm{x}_i\in\mathbb{R}^D  \}_{1\le i\le m}$ with $d < D$ via maximizing their variances in the low-dimensional subspaces \cite{2010pca}. Specifically for
 $d=1$, (linear) PCA obtains $\boldsymbol{\chi}_i:=\bm{u}^\top\bm{x}_i$, where the direction $\bm{u}\in\mathbb{R}^D$ is found by
 \begin{subequations}
 	\label{eq:pca}
 	 \begin{align}
 	 \underset{\bm{u}\in\mathbb{R}^{D}}{\max}
 	 ~~~\,&\bm{u}^\top\bm{C}_{xx}\bm{u}\\
 	 {\rm s.\;to}~~~~ &\bm{u}^\top\bm{u}=1\label{eq:pca2}.
 	 \end{align}
 \end{subequations}
Solving \eqref{eq:pca} yields $\bm{u}$ 
 as the principal eigenvector of matrix $\bm{C}_{xx}$, also known as the first PC. Instead of having constraint \eqref{eq:pca2} explicitly, we assume wlog that the solution $\bm{u}$ will always be normalized to have unity norm. 
For $d> 1$, PCA amounts to computing the first $d$ principal eigenvectors of $\bm{C}_{xx}$ instead.  
As alluded to in Sec. \ref{sec:intro}, when two datasets $\{\bm{x}_i \}$ and $\{\bm{y}_j\}$ are available, PCA performed either on  $\{\bm{x}_i\}$, or on $\{\{\bm{x}_i \}, \{\bm{y}_j \}\}$, can generally not unveil the patterns or trends that are specific to the target relative to the control data.


Contrastive (c) PCA \cite{2017cpca}, on the other hand, aims to identify directions $\bm{u}$ along which the target data possesses large variations while the control data has small variations. 
Concretely,
cPCA pursues problem \cite{2017cpca} 
 	\begin{align}
 	&\underset{\|\bm{u}\|_2=1}{\max}
 	~~~\,\bm{u}^\top\bm{C}_{xx}\bm{u}-\alpha \bm{u}^\top\bm{C}_{yy}\bm{u}.\label{eq:cpca}
 	\end{align}
 whose solution 
is given by the eigenvector of $\bm{C}_{xx}-\alpha\bm{C}_{yy}$ associated with the largest eigenvalue, and constitutes the first contrastive (c) PC.
 Here, $\alpha>0$ is a hyper-parameter that trades off maximizing the target data variance (the first term in \eqref{eq:cpca}) for minimizing the control data variance (second term). 
 However, there is no 
  rule of thumb for choosing $\alpha$. Although a spectral clustering based algorithm has been developed to automatically select the value of $\alpha$, its brute-force search
discourages 
  its use in large-scale datasets \cite{spectralclustering}.


\vspace{-0pt}
\section{The novel approach}
\label{sec:dpca}


Unlike PCA, LDA is ``supervised,'' and seeks directions that yield the largest possible separation between classes 
 via maximizing the ratio of the variance across classes to the variance within classes. In the same vein, when both the target and the background data are available, and one is interested in extracting features, namely component vectors that are \emph{only} present in the target data but \emph{not} in the background data,
 a meaningful approach would be to maximize the ratio of the variance of the target data over that of the background data
 \begin{align}
 \underset{\|\bm{u}\|_2=1}{\max}
 ~~~\,\frac{\bm{u}^\top\bm{C}_{xx}\bm{u}}{ \bm{u}^\top\bm{C}_{yy}\bm{u}}\label{eq:dpca0}
 \end{align}
 which, with slight abuse of the term ``discriminant,'' we call \emph{discriminative PCA}. Likewise, the solution of \eqref{eq:dpca0} will be termed first discriminative (d) PC, or dPC for short. 

\subsection{dPCA Algorithm}
Suppose that 
$\bm{C}_{yy}$ is full rank 
with eigen-decomposition
 $\bm{C}_{yy}:=\bm{U}_y^\top\bm{\Sigma}_y\bm{U}_y$. Upon defining $\bm{C}_{yy}^{1/2}:=\bm{\Sigma}_y^{1/2}\bm{U}_y$,
and changing variables $\bm{v}:=\bm{C}_{yy}^{1/2}\bm{u}$, \eqref{eq:dpca0} admits the same solution as
 	\begin{equation}
 	\bm{v}^\ast:=\arg\underset{
 		\|\bm{v}\|_2=1
 		}{\max}
 	~~~\,\bm{v}^\top\bm{C}^{-\top/2}_{yy}\bm{C}_{xx}\bm{C}^{-1/2}_{yy}\bm{v}\label{eq:dpca}
 	\end{equation}
 which is the principal eigenvector of $\bm{C}_{yy}^{-\top/2}\bm{C}_{xx} \bm{C}_{yy}^{-1/2}$. Finally, the solution to \eqref{eq:dpca0} is recovered as $\bm{u}^\ast:=\bm{C}_{yy}^{-1/2}\bm{v}^\ast$, followed by normalization to obtain a unit norm.

  On the other hand, leveraging Lagrangian duality, the solution of \eqref{eq:dpca0} can also be obtained as the right eigenvector of $\bm{C}_{yy}^{-1}\bm{C}_{xx}$. To see this, note that \eqref{eq:dpca0} can be
  rewritten as 
  \begin{subequations}
  	\label{eq:const}
  	\begin{align}
  		\underset{\bm{u}\in\mathbb{R}^{D}}{\max}
  		~~~\,&\bm{u}^\top\bm{C}_{xx}\bm{u}\label{eq:const1}\\
  		{\rm s.~to}~~~~&\bm{u}^\top\bm{C}_{yy}\bm{u}=b\label{eq:const2}
  	\end{align}
  \end{subequations}
   for some constant $b>0$ such that the solution $\|\bm{u}^\ast\|_2=1$. One can simply set $b=1$ and subsequently normalize the solution of  \eqref{eq:const}. Letting $\lambda\in\mathbb{R}$ be the dual variable corresponding to constraint \eqref{eq:const2}, the Lagrangian of \eqref{eq:const} is
   \begin{equation}
   	\label{eq:lag}
   	\mathcal{L}(\bm{u};\lambda)=\bm{u}^\top\bm{C}_{xx}\bm{u}+\lambda\!\left(1-\bm{u}^\top\bm{C}_{yy}\bm{u}\right).
   \end{equation}    
   
   The KKT conditions assert that for the optimal  $(\bm{u}^\ast;\lambda^\ast)$, it holds that $\bm{C}_{xx}\bm{u}^\ast=\lambda^\ast\bm{C}_{yy}\bm{u}^\ast$, which is a generalized eigenvalue problem. Equivalently, one can rewrite
   \begin{equation}
   	\bm{C}_{yy}^{-1}\bm{C}_{xx}\bm{u}^\ast=\lambda^\ast\bm{u}^\ast
   \end{equation}
   suggesting that $\bm{u}^\ast$ is an eigenvector of  $\bm{C}_{yy}^{-1}\bm{C}_{xx}$ associated with eigenvalue $\lambda^\ast$. Respecting the constraint $(\bm{u}^\ast)^\top\bm{C}_{yy}\bm{u}^\ast=1$, the objective \eqref{eq:const1} reduces to 
   \begin{equation}
   (\bm{u}^\ast)^\top\bm{C}_{xx}\bm{u}^\ast=\lambda^\ast(\bm{u}^\ast)^\top\bm{C}_{yy}\bm{u}^\ast=\lambda^\ast.
   \end{equation}
   It is clear now that the optimal objective value of problem \eqref{eq:const} is equal to the largest eigenvalue of $\bm{C}_{yy}^{-1}\bm{C}_{xx}$, and the optimal solution $\bm{u}^\ast$ is the corresponding eigenvector.
   
    \begin{algorithm}[t]
   	\caption{Discriminative principal component analysis.}
   	\label{alg:dpca}
   	\begin{algorithmic}[1]
   		\STATE {\bfseries Input:}
   		Nonzero-mean target and background data $\{\accentset{\circ}{\bm{x}}_i\}_{1\le i\le m}$, $\{\accentset{\circ}{\bm{y}}_j\}_{1\le j\le n}$; number of dPCs $d$.
   		\STATE {\bfseries Remove} the mean from $\{\accentset{\circ}{\bm{x}}_i\}$ and $\{\accentset{\circ}{\bm{y}}_j\}$ to yield centered data $\{\bm{x}_i\}$, and $\{\bm{y}_j\}$.
   		\STATE {\bfseries Construct} the sample covariance matrices:
   		\begin{equation*}
   		   		\vspace{-4pt}
   		\bm{C}_{xx}:=\frac{1}{m}\sum_{i=1}^m\bm{x}_i\bm{x}_i^\top,\!\quad {\rm and}\quad \!
   		\bm{C}_{yy}:=\frac{1}{n}\sum_{j=1}^n\bm{y}_j\bm{y}_j^\top
   		   		\vspace{-4pt}
   		\end{equation*}
   		\STATE {\bfseries Perform} \label{step:4} SVD
   		on matrix $\bm{C}_{yy}^{-1}\bm{C}_{xx}$.
   		\STATE {\bfseries Output}\label{step:5} 
   		the $d$ (right) singular vectors corresponding to the $d$ largest singular values. 

   	\end{algorithmic}
   \end{algorithm}

  For $d>1$, one finds the $d$ (right) eigenvectors of $\bm{C}_{yy}^{-1}\bm{C}_{xx}$ that correspond to the $d$ largest eigenvalues as
  the first $d$ dPCs. 
  For ease of implementation, the proposed dPCA approach for contrastive data exploration is summarized in Alg. \ref{alg:dpca}. 
Concerning dPCA, a couple of remarks are in order.
   		\vspace{-5pt}
 \begin{enumerate}[wide,labelwidth=!,labelindent=0pt,topsep=1pt,parsep=1pt,itemsep=2pt]
 	\item[\textbf{Remark 1.}] 
 	When there is no background data, with $\bm{C}_{yy}=\bm{I}_D$, dPCA boils down to PCA. 
 	On the other hand, when there are multiple background datasets, one can first combine them into a single one, and then apply dPCA. Other twists will be discussed in the full version of this paper. 
 	\item[\textbf{Remark 2.}] 
 	Performing dPCA on  $\{\bm{x}_i\}$ and $\{\bm{y}_j\}$ can be seen as performing PCA on the transformed data $\{\bm{C}^{-\top/2}_{yy}\bm{x}_i \}$ to yield $\bm{v}^\ast$, followed by a linear re-transformation $\bm{u}^\ast=\bm{C}_{yy}^{-1/2}\bm{v}^\ast$. The new data can be understood as the data obtained after removing the ``background" component vectors 
 	 from the target data.
 	 	\item[\textbf{Remark 3.}] Inexpensive power or Lanczos iterations \cite{saad1} can be employed to compute the principal eigenvectors in \eqref{eq:dpca}.
 \end{enumerate}


\vspace{-6pt}
 \subsection{dPCA vis-{\` a}-vis cPCA}
 
 Consider again the constrained form of dPCA  \eqref{eq:const} and its Lagrangian \eqref{eq:lag}. Using Lagrange duality, when choosing $\alpha=\lambda^\ast$ in \eqref{eq:cpca}, cPCA maximizing
$\bm{u}^\ast(\bm{C}_{xx}-\lambda^\ast\bm{C}_{yy})\bm{u}$
 is equivalent to 
 $\max_{\bm{u}\in\mathbb{R}^{D}}\mathcal{L}(\bm{u};\lambda^\ast)=\bm{u}^\top\left(\bm{C}_{xx}-\lambda^\ast\bm{C}_{yy}\right)\bm{u}+\lambda^\ast$, 
 which is exactly dPCA. In other words, cPCA and dPCA are equivalent when $\alpha$ in cPCA is carefully set as the optimal dual variable $\lambda^\ast$ for the constrained form \eqref{eq:const} of dPCA, namely the largest eigenvalue of $\bm{C}_{yy}^{-1}\bm{C}_{xx}$. 
 
It will be useful for further analysis to focus on simultaneously diagonalizable matrices $\bm{C}_{xx}$ and $\bm{C}_{yy}$, that is
\begin{equation}
	\label{eq:diag}
	\bm{C}_{xx}:=\bm{U}^\top\bm{\Sigma}_x\bm{U},~~~{\rm and}~~~\,\bm{C}_{yy}:=\bm{U}^\top\bm{\Sigma}_y\bm{U}
\end{equation} 
where $\bm{U}\in\mathbb{R}^{D\times D}$ is unitary and simultaneously decomposes $\bm{C}_{xx}$ and $\bm{C}_{yy}$, while diagonal matrices $\bm{\Sigma}_x,\,\bm{\Sigma}_y\succ \bm{0}$ hold accordingly eigenvalues $\{\lambda_x^i\}_{1\le i\le D}$ of $\bm{C}_{xx}$ and $\{\lambda_y^i\}_{1\le i\le D}$ of $\bm{C}_{yy}$ on their main diagonals. It clearly holds that $\bm{C}_{yy}^{-1}\bm{C}_{xx}=\bm{U}^\top\bm{\Sigma}_y^{-1}\bm{\Sigma}_x\bm{U}=\bm{U}^\top {\rm diag}\big(\{\frac{\lambda_x^i}{\lambda_y^i}\}_{1\le i\le D}\big)\bm{U}$. Looking for the first $d$ dPCs boils down to 
taking the $d$ columns of $\bm{U}$ 
associated with the $d$ largest eigenvalue ratios among $\{\frac{\lambda_x^i}{\lambda_y^i}\}_{1\le i\le D}$.

On the other hand, the solution of cPCA under a given $\alpha$, or the first $d$ cPCs of $\bm{C}_{xx}-\alpha\bm{C}_{yy}=\bm{U}^\top(\bm{\Sigma}_x-\alpha\bm{\Sigma_y})\bm{U}=\bm{U}^\top{\rm diag}\big(\{\lambda_x^i-\alpha\lambda_y^i\}_{1\le i \le D}\big)\bm{U}$, are found as 
the $d$ columns of $\bm{U}$ that correspond to the $d$ largest numbers in $\{\lambda_x^i-\alpha\lambda_y^i\}_{1\le i \le D}$.
In the ensuing section, we show that when given data obey certain models, dPCA is LS optimal. 

\section{Optimality of \MakeLowercase{d}PCA}

Adopting a bilinear (factor analysis) model, PCA describes the (non-centered) data $\{\accentset{\circ}{\bm{y}}_j\}_{1\le j\le n}$ as 
\begin{equation}
\accentset{\circ}{\bm{y}}_j=\bm{m}_y+
\bm{U}_y\bm{\psi}_j+\bm{e}_{y,j},	\quad\; 1\le j\le n\label{eq:y}	
\end{equation}
where 
$\bm{m}_y$ is a location vector, 
$\bm{U}_y\in\mathbb{R}^{D\times D}$ has orthonormal columns; 
$\{\bm{\psi}_{j}\}_{1\le j\le n}$ are the coefficients, and
$\{\bm{e}_{y,j}\}_{1\le j\le n}$ zero-mean random variables. The unknowns $\bm{m}_y$, $\bm{U}_y$, and $\{\bm{\psi}_{i}\}_{1\le j\le n}$ can be estimated using the LS criterion as \cite{1995pca}
\begin{equation}
\underset{\bm{m}_y,\;\{\bm{\psi}_j\},\atop\bm{U}_y^\top\bm{U}_y=\bm{I}}{\minimize} ~~~\sum_{j=1}^n\left\|\accentset{\circ}{\bm{y}}_j-\bm{m}_y-\bm{U}_y\bm{\psi}_j\right\|_2^2.\label{eq:lsy}
\end{equation}
whose solution is provided as \cite{1995pca}:
$\bm{m}_y^\ast:=(1/n)\sum_{j=1}^n{\accentset{\circ}{\bm{y}}}_j$, $\bm{\psi}_j^\ast:=(\bm{U}_y^\ast)^\top(\accentset{\circ}{\bm{y}}_j-\hat{\bm{m}}_y)$, $\forall 1\le j\le n$, and $\bm{U}_y^\ast$ stacks up as its columns the eigenvectors of
$\bm{C}_{yy}:=(1/n)\sum_{j=1}^n\bm{y}_j\bm{y}_j^\top$, to form $\bm{C}_{yy}:=\bm{U}_y^\ast\bm{\Sigma}_y(\bm{U}_y^\ast)^\top$,
where $\bm{y}_j:={\accentset{\circ}{\bm{y}}}_j-\bm{m}_y^\ast$ is the centered data. For notational brevity, the superscript ${}^\ast$ shall be dropped when clear from the context.
Wlog, let $\bm{U}_y:=[\bm{U}_b~\bm{U}_n]$ be partitioned such that $\bm{U}_b\in\mathbb{R}^{D\times k}$ corresponds to the first $k$ PCs of $\bm{C}_{yy}$, which capture most background component vectors.

In the context of discriminative data analysis, we \emph{assume} that the target data share some PCs with the background data (say $\bm{U}_b$ of \eqref{eq:y}), and has additionally a few (say $d$) PCs that capture patterns specific to the target data but are less significant than the PCs in $\bm{U}_b$.
For simplicity, consider $d=1$, and model $\{\accentset{\circ}{\bm{x}}_i\}$ as
\begin{equation}
\accentset{\circ}{\bm{x}}_i=\bm{m}_x+\left[\bm{U}_b~\;\bm{u}_s\right]\left[\!\!\begin{array}{c}\boldsymbol{\chi}_{b,i}\\{\chi}_{s,i}\end{array}\!\!\right]+\bm{e}_{x,i},\quad 1\le i\le m\label{eq:x}
\end{equation}
where $\bm{m}_x$ is the mean of $\{\accentset{\circ}{\bm{x}}_i\}_{1\le i\le m}$; and assuming $k+d\le D$, $\bm{U}_x:=[\bm{U}_b~\;\bm{u}_s]\in\mathbb{R}^{D\times (k+1)}$ has orthonormal columns, where $\bm{U}_b$ describes the component vectors present both in the background as well as target data, while $\bm{u}_s\in\mathbb{R}^{D\times 1}$ captures the patterns of interest that are present \emph{only} in the target data. 
Our goal is to obtain this discriminative subspace $\bm{U}_s$ given solely the two datasets. By modeling this distinctly informative component $\chi_{s,i}\bm{u}_s$ in \eqref{eq:x} explicitly as an outlier vector, it is also possible to employ robust PCA which boils down to solving a nonconvex optimization problem \cite{2012rpca}.

Likewise, remove the mean $\bm{m}_x:=(1/n)\sum_{i=1}^n\accentset{\circ}{\bm{x}}_i$ from the target data yielding $\bm{x}_i:=\accentset{\circ}{\bm{x}}_i-\bm{m}_x$. Recalling $\bm{C}_{yy}^{1/2}:=\bm{\Sigma}_y^{1/2}\bm{U}_y$,
 consider the transformed data model for $ 1\le i\le m$:
\begin{align}
\!\!	\bm{C}_{yy}^{\top/2}\bm{x}_i&=		\bm{C}_{yy}^{\top/2}\left[\bm{U}_b~\;\bm{u}_s\right]\left[\!\!\begin{array}{c}\boldsymbol{\chi}_{b,i}\\{\chi}_{s,i}\end{array}\!\!\right]+	\bm{C}_{yy}^{\top/2}\bm{e}_{x,i}\nonumber\\
	&=	\chi_{s,i}\bm{C}_{yy}^{\top/2}\bm{u}_s+	\bm{C}_{yy}^{\top/2}\bm{e}_{x,i}:={\chi}_{s,i}\tilde{\bm{u}}_s+\tilde{\bm{e}}_{x,i}
	\label{eq:tx}
\end{align}
where $	\bm{C}_{yy}^{\top/2}\bm{U}_b$ vanishes due to the orthogonality of columns of $\bm{U}_y^\ast=[\bm{U}_b~\bm{U}_n]$, $\tilde{\bm{u}}_s:=(\bm{U}_y^\ast)^\top\bm{u}_s$, 
and $\tilde{\bm{e}}_{x,i}$ is a zero-mean random variable.
Similar to \eqref{eq:lsy}, the LS optimal estimate $\tilde{\bm{u}}_s^\ast$ is given by 
the first principal eigenvector of $$\tilde{\bm{C}}_{xx}:=\frac{1}{m}\sum_{i=1}^m\bm{C}_{yy}^{\top/2}\bm{x}_i\bm{x}_i^\top\bm{C}_{yy}^{\top/2}=\bm{C}_{yy}^{\top/2}\bm{C}_{xx}\bm{C}_{yy}^{1/2}.$$ 
Hence, the discriminative PCs can be recovered from $\tilde{\bm{u}}_s$ as $\bm{u}_s^\ast:=\bm{C}_{yy}^{1/2}\tilde{\bm{u}}_s^\ast$, which coincides with solutions of problem \eqref{eq:dpca0} or \eqref{eq:dpca}, and establishes the LS optimality of dPCA.

  

%


\begin{figure}[t]
	\centering 
	\includegraphics[scale=0.54]{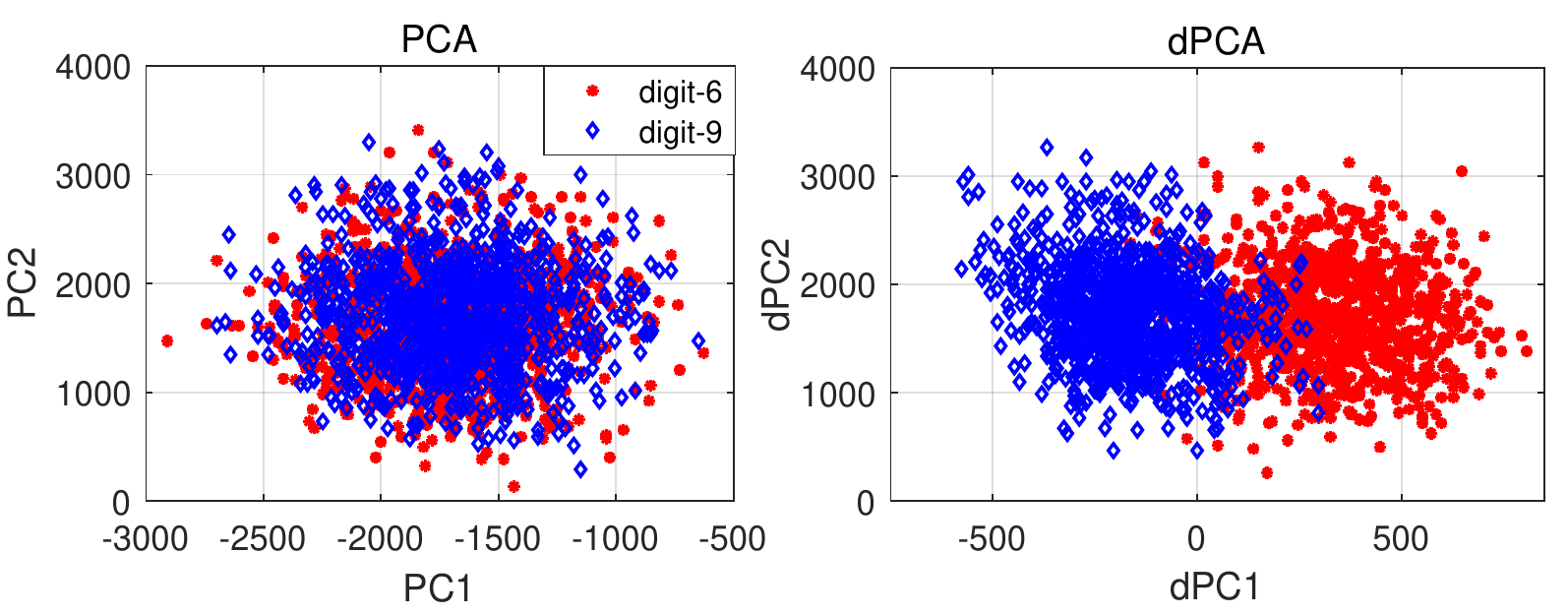} 
	\vspace{-20pt}
	\caption{\small{dPCA versus PCA on semi-synthetic data.}}
	\vspace{-10pt}
	\label{fig:digits}
\end{figure}

	\vspace{-2pt}
\section{Numerical Tests}
		\vspace{-2pt}
In this section, the performance of dPCA is assessed relative to PCA and cPCA \cite{2017cpca} on a synthetic and a real dataset.
In the first experiment, (semi-)synthetic target and background data were generated from real images. Specifically, the target data were constructed using $2,000$ handwritten digits $6$ and $9$ ($1,000$ for each) of size $28 \times 28$ from the MNIST dataset \cite{mnist} superimposed with $2,000$ frog images from the CIFAR-$10$ dataset \cite{cifar10}. The raw $32\times32$ frog images were converted to grayscale and randomly cropped to $28 \times 28$. 
The background data were built with $3,000$ resized images only, which were sampled randomly from the remaining frog images.


\begin{figure}[t]
	\centering 
	\includegraphics[scale=0.62]{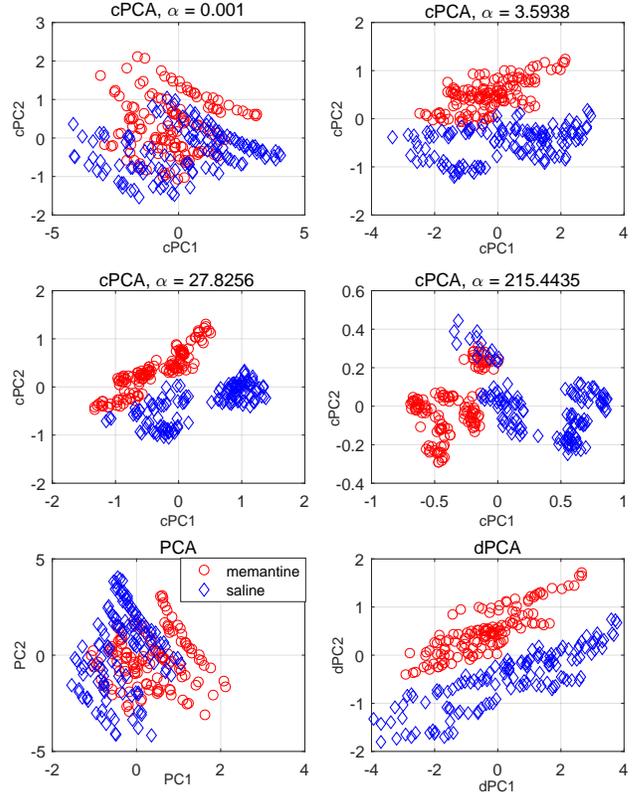} 
	\vspace{-12pt}
	\caption{\small{Discovering subgroups in mice protein expression data.}}
	\label{fig:mice}
	\vspace{-12pt}
\end{figure}

We performed PCA on the target data only. The results of the target images embedded onto the first two PCs and two dPCs are depicted in the left and right panels of Fig. \ref{fig:digits}, respectively. 
Clearly, PCA is unable to discover the two digit subgroups. This is because 
the obtained two PCs are likely associated with the background component vectors within the target images, namely features depicting frog images rather than handwritten digits. 
On the contrary, two clusters emerged in the plot of dPCA, demonstrating its efficacy over PCA in discriminating unique features of one dataset from the other.     

The capability of dPCA in discovering subgroups is further tested on real  protein expression data. In the second experiment, the target data consist of $267$ points, each recording $77$ protein expression measurements for a mouse suffering Down Syndrome \cite{mice}. There were $135$ mice with drug-memantine treatment as well as $134$ without treatment. The control data comprise such measurements from $135$ healthy mice. The $135$ control mice are likely to exhibit similar natural variations (due to e.g., sex and age) as the target mice, but without the differences that result from Down Syndrome. 
For cPCA, the spectral-clustering algorithm in \cite{spectralclustering} was implemented for 
 selecting $4$ from a list of $15$ logarithmically spaced values between $0.001$ and $1,000$ \cite{2017cpca}. The simulated results are presented in Fig. \ref{fig:mice}, with red circles denoting mice with treatment and blue diamonds the other mice.
 PCA reveals that the two types of mice follow a similar distribution in the space spanned by the first two PCs; see the left bottom plot in Fig. \ref{fig:mice}. The separation between the two groups of mice becomes clear when dPCA is applied. At the price of runtime ($15$ times more than dPCA), 
 cPCA with properly \emph{learnt} parameters ($\alpha=3.5938$ and $27.8256$) can work well too.
  

 		\vspace{-7pt}
\section{Conclusions}
		\vspace{-5pt}
This paper advocated a novel approach termed dPCA for discriminative analytics, namely for discovering the most informative features that are specific to one dataset but are also distinct from some other correlated datasets. The resultant algorithm 
amounts to 
solving a generalized eigenvalue problem. Comparing to existing alternatives,
dPCA bypasses parameter tuning, and incurs complexity required to perform only one SVD. It is provably optimal in the LS sense provided that the background component vectors are present in the target data. Simulated tests using (semi)-synthetic images and real protein expression data corroborated the merits of the developed approach. Investigating dPCA 
using kernels and over graphs constitutes meaningful future research directions. 


%


\newpage

\bibliographystyle{IEEEtran}
\bibliography{pca}

\end{document}